\documentclass[11pt,a4paper]{article}

\usepackage[utf8]{inputenc}
\usepackage[T1]{fontenc}
\usepackage{mathpazo}          
\usepackage{amsmath,amssymb}
\usepackage{graphicx}
\usepackage{booktabs}
\usepackage{array}
\usepackage{tabularx}
\usepackage[margin=1in]{geometry}
\usepackage{setspace}
\usepackage{natbib}
\usepackage{url}
\usepackage[hidelinks]{hyperref}
\usepackage{xcolor}
\usepackage{float}
\usepackage{enumitem}
\usepackage{fancyhdr}
\usepackage{caption}
\usepackage{tikz}
\usetikzlibrary{arrows.meta,positioning}
\usepackage{tocloft}
\newcommand{\doi}[1]{\href{https://doi.org/#1}{doi:#1}}

\title{\textbf{Prompt Design at Scale:}\\[6pt]
How Format, Instruction Count, and Context Length Shape\\
Instruction Adherence and Hallucination in Large Language Models}

\author{Netanel Eliav\\[4pt]
Machine Human Intelligence Lab (MHIL)\\
\texttt{netanel@mhil.org} \quad \texttt{inetanel@me.com}\\
\url{https://mhil.org}}

\date{July 2026\\[6pt]
\small Preprint --- submitted to arXiv cs.AI, cs.CL\\
\small This paper has not undergone peer review.}

\begin{document}
\maketitle
\thispagestyle{fancy}

\begin{abstract}
Practitioners make three prompt-design decisions with almost no controlled evidence behind any of them: how to format instructions and injected context (markdown, plain text, prose, or tabular), how many simultaneous instructions a system prompt can carry before compliance degrades, and how much context a model can hold before recall and honesty degrade. We report two controlled experiments crossing all three factors on one held, contamination-free synthetic corpus (the ``Book of Veyra,'' 8,780 uniquely-named entities, deterministically regenerable from a fixed seed), evaluated across five models spanning two capability tiers and three model families (Claude Sonnet~5, Claude Haiku, Gemini Flash, and two Qwen open-weight sizes).

Experiment~1 (960 calls/model) measures instruction-following decay as rule count $N$ grows from 10 to 160, crossed with four rendering formats and system-prompt vs.\ user-turn placement. Perfect-response rate collapses to zero by $N=80$ for every model, every format, and both placements, a floor effect essentially independent of format. Placement produces effects at least as large as format at $N=160$ in four of five models, but the direction is model-specific (it helps two models, hurts two, and has no effect for one). No model shows a reliable markdown advantage, and one 35B open-weight model reliably favors plain text instead.

Experiment~2 (5,520 calls/model at full context) measures recall accuracy, false-premise sycophancy, and absent-fact fabrication across a 2k-to-512k-token context ladder in the same four formats. Recall stays near ceiling through roughly 64--128k tokens, then degrades sharply and format-dependently: one model's accuracy spread across formats reaches 48 percentage points at 128k tokens, and the two models tested to 512k show format-accuracy spreads that scale with proximity to each model's own effective context ceiling rather than with absolute token count. Fabrication never occurs (0/5,760 absent-fact probes across all models, rungs, and formats), and sycophancy stays negligible ($\leq 8.3\%$) throughout. What actually rises sharply near each model's own ceiling is outright refusal to answer (0\% to 79--90\%), a distinct failure mode from either sycophancy or fabrication. Neither of two pre-registered format orderings holds: the same format is the best performer for one model/rung and the worst for another. Accounting for format's measured token overhead (+22\% to +37\% over plain text) further changes which format is preferable in the cases with genuine accuracy spread to adjust.

We release the full harness, corpus generator, and raw results (VeyraBench) for exact, byte-identical reproduction.\footnote{\url{https://github.com/iNetanel/veyrabench}}
\end{abstract}

\noindent\textbf{arXiv categories:} cs.AI (primary); cs.CL (cross-listed)\\[8pt]

\tableofcontents
\newpage

\section{Introduction}

Consider two routine moments in how practitioners build LLM-backed systems in 2026. First, an engineer writing a system prompt accumulates thirty, then eighty, then a hundred and sixty simultaneous formatting rules, tone constraints, and safety instructions, and assumes the model will keep pace as long as each rule is phrased clearly. Second, a retrieval-augmented pipeline feeds a growing document store, two thousand tokens today and five hundred thousand tomorrow, into a single context window, rendered however the ingestion script happens to format it, and the team trusts the model to find, remember, and honestly report what is actually there. Both moments involve a decision about \emph{format}: markdown headers and bullets, flat plain text, flowing prose, or a table. But both are at least as much a decision about \emph{scale}: how much is being asked of the model at once, and how much it is being asked to hold in mind. Practitioners argue constantly about the first question. The second is usually assumed away.

This paper treats that as a mistake and argues that format cannot be evaluated in isolation from scale. A markdown bullet list and a plain-text sentence may behave identically at ten instructions and diverge sharply at a hundred. A table and a paragraph may return identical answers at two thousand tokens of context and diverge sharply at five hundred thousand. Existing controlled work studies format sensitivity while holding scale roughly fixed \citep{Sclar2024,He2024}, or studies scale (growing instruction counts, growing context length) largely without varying format \citep{Zhou2023,Jaroslawicz2025,Liu2023,Hong2025}. No publicly available study, to our knowledge, crosses format against both scale axes, instruction count and context length, on one held, controlled corpus. This paper does.

We report two experiments that share a single design principle. We render the same underlying content in four formats (markdown, plain text, prose, and a markdown table) and push each format along its own natural scale axis until it breaks.

\textbf{Experiment 1 (instructions).} A system prompt or user turn carries $N \in \{10, 20, 40, 80, 120, 160\}$ simultaneous, programmatically verifiable rules (word-count ranges, required and forbidden words, exact opening/closing text, paragraph counts), rendered in each of the four formats, with the rule block placed either in the system prompt or the user turn. We measure the resulting instruction-following \emph{decay curve}, how compliance falls as $N$ grows, separately by format, by placement, and by model.

\textbf{Experiment 2 (context).} A deterministic, contamination-free 512,000-token synthetic corpus (the Book of Veyra: 8,780 uniquely named fictional entities with interlocking, overlapping attributes, generated with no LLM involvement from a fixed seed) is rendered in the same four formats and sliced into a context ladder from 2,000 to 512,000 tokens. At each rung we probe recall accuracy, sycophantic agreement with false premises, and fabrication of facts never stated, separately by format and by context length.

Both experiments run against five models spanning two capability tiers within one provider (Claude Sonnet~5, Claude Haiku) and one additional provider/architecture point (Gemini Flash), plus a within-family open-weight size comparison (Qwen 27B, Qwen 35B). Table~\ref{tab:models} reports the exact model identifier queried for each. Section~\ref{sec:related} positions this design against the closest prior work on format sensitivity, instruction-count scaling, structured-vs-unstructured context, long-context sycophancy, and synthetic hallucination benchmarks. Several of these works independently anticipate pieces of what we find, and we engage with them directly rather than eliding the overlap.

\begin{table}[t]
\centering
\caption{Models evaluated: display name used in this paper vs.\ the exact identifier queried}
\label{tab:models}
\small
\begin{tabular}{llll}
\toprule
\textbf{Display name} & \textbf{Model identifier} & \textbf{Host} & \textbf{Context ceiling used}\textsuperscript{*} \\
\midrule
Claude Sonnet 5 & \texttt{claude-sonnet-5} & Anthropic API & 1{,}000{,}000 tokens \\
Claude Haiku    & \texttt{claude-haiku-4-5} & Anthropic API & 200{,}000 tokens \\
Gemini Flash    & \texttt{gemini-3.5-flash} & Google (OpenAI-compatible endpoint) & 1{,}000{,}000 tokens \\
Qwen 27B        & \texttt{qwen\_27b} & Self-hosted (Ollama) & 220{,}000 tokens \\
Qwen 35B        & \texttt{qwen\_35b} & Self-hosted (Ollama) & 180{,}000 tokens \\
\bottomrule
\end{tabular}

\vspace{4pt}
\footnotesize \textsuperscript{*}The configured ceiling used to decide which context-ladder rungs a model is evaluated at in Experiment~2 (Section~\ref{sec:exp2}), after applying a measured real-token-to-estimate multiplier per model (1.1--1.6$\times$, provider-dependent) as a safety margin against the model's advertised window; it is not necessarily each vendor's own stated maximum. Both Qwen models are served with extended-thinking/reasoning generation disabled at the API level for both experiments. All models are sampled at their own provider default temperature (no temperature override), with 3 repeats per question in Experiment~2 and 20 trials per cell in Experiment~1.
\end{table}

\subsection{Contributions}

This paper makes the following contributions:

\begin{enumerate}[leftmargin=*]
\item \textbf{A joint decay-and-placement analysis of instruction-following.} We compare adherence-vs-$N$ decay curves across four formats and two prompt placements (system vs.\ user turn) on one instruction pool. Threshold, or ``knee,'' decay in instruction count has been independently documented at larger scale by concurrent work \citep{Jaroslawicz2025,Harada2025}. Our contribution is narrower and more specific: we cross that decay curve against format \emph{and} against placement, neither of which prior work varies. We find placement produces effects comparable to or larger than format, in a direction that is itself model-specific.
\item \textbf{A four-format, four-model, six-rung factorial for long-context hallucination.} Recall error, sycophancy, and fabrication are measured jointly across a 2k--512k-token ladder in four formats on a single contamination-free corpus. We are not aware of a prior study that crosses format with context length for all three failure modes simultaneously.
\item \textbf{A structure-vs-syntax decomposition.} By holding content byte-identical across a list-structured markdown format, a table format, an unstructured plain format, and unstructured prose, we separate whether \emph{any} structure helps from whether \emph{tabular} structure specifically helps. Recent work independently shows tables outperform unstructured text on other corpora \citep{Oh2024,Liu2025format}. Our contribution is a tighter four-way decomposition on one held corpus that additionally crosses this question with context length, which that work does not.
\item \textbf{Token-overhead-adjusted accuracy.} Using real subword token counts, we show format's accuracy advantages do not always survive its cost. In the clearest test cases, converting to accuracy-per-token either reverses an apparent ranking or reinforces an already-cheaper format's win.
\item \textbf{VeyraBench:} the full benchmark harness, the Book of Veyra corpus generator, and all raw model outputs, released for byte-identical reproduction from a fixed seed. The corpus itself follows an active line of work on synthetic, leakage-resistant benchmark construction \citep{Bang2025,Li2024needlebench}. Our specific contribution is a single corpus rendered in four parallel, content-identical formats, purpose-built to isolate format effects rather than to test recall alone.
\end{enumerate}

\subsection{Paper Structure}

Section~\ref{sec:related} reviews related work on format sensitivity, instruction-count scaling, structured context, long-context sycophancy, and synthetic benchmarks, and positions our contributions against it directly. Section~3 describes the Book of Veyra corpus and its four renderings. Sections~4 and~5 present Experiment~1 and Experiment~2. Section~6 presents the structure-vs-syntax decomposition. Section~7 presents the cost-adjusted accuracy analysis. Section~8 distills practical guidance. Section~9 states limitations. Section~10 outlines future work. Section~11 concludes.

\section{Related Work}
\label{sec:related}

Our design sits at the intersection of five literatures that have, until now, developed largely independently: prompt format sensitivity, instruction-count scaling, structured-versus-unstructured context, long-context sycophancy, and synthetic contamination-free benchmarks. Each is reviewed in turn, with an explicit statement of how our design differs.

\subsection{Prompt Format Sensitivity}

\citet{Sclar2024} show that semantically equivalent prompts differing only in superficial formatting (spacing, separators, casing) produce accuracy swings of up to 76 points on some models, establishing that format is not a cosmetic concern. \citet{He2024} directly compare plain text, markdown, JSON, and YAML system-prompt formatting on GPT-3.5 and GPT-4 and find no universally best format, with format sensitivity decreasing but not disappearing at higher capability. Both studies hold task complexity roughly fixed while varying format. Neither varies instruction count or context length alongside format, which is the axis this paper adds.

\subsection{Instruction-Count Scaling}

A separate literature studies how compliance degrades as the number of simultaneous instructions grows, without varying prompt format. \citet{Zhou2023} introduce a verifiable-instruction methodology (IFEval) that this paper's Experiment~1 rule design borrows. \citet{Jaroslawicz2025} scale instruction density from 10 to 500 keyword-inclusion rules across 20 models and 7 providers, and identify a ``threshold decay'' pattern, stable performance followed by a sharper decline past a knee, for the strongest models in their pool. \citet{Harada2025} fit a closed-form relationship between prompt-level accuracy and per-instruction accuracy on up to ten simultaneous instructions across five model families. Both studies independently establish decay-with-instruction-count as a robust, now multiply-replicated phenomenon, but neither varies prompt format as a factor. Our Experiment~1 measures the same underlying decay phenomenon but crosses it against format and against system-vs-user-turn placement, on a narrower instruction-count range (10--160) than \citet{Jaroslawicz2025}'s work but with both of those additional factors present.

\subsection{Long-Context Retrieval and Sycophancy}

\citet{Liu2023} document a U-shaped retrieval-accuracy curve by information position across six model families, and \citet{Hong2025} extend degradation measurement to 18 production models across a 10,000--500,000-token range, finding no model retrieves reliably across its full advertised window. Neither study varies the format of the injected context. On sycophancy specifically, recent work manipulates conversational or memory context rather than document format. \citet{Jain2026} show agreement with a user's prior stated views rises with accumulated conversational and memory context across 38 real users. \citet{Prasad2026}, in a large non-peer-reviewed trial set (80,433 trials, 6 models), finds that raw conversational-context length has an architecture-dependent effect on sycophancy, reliably present only below roughly 20--24B active parameters. The \emph{composition} of that context (agreement- vs.\ correction-leaning filler) is a substantially larger effect than its length. Our Experiment~2 measures sycophancy, and separately outright fabrication and simple recall error, as a joint function of single-document context length \emph{and} format, which neither line of prior work does.

\subsection{Structured versus Unstructured Context}

Whether structuring injected data as a table specifically, rather than any structured form, improves factual accuracy has begun to receive direct attention. \citet{Oh2024} compare natural text, structured text, JSON, knowledge graphs, and tables for factual-data interaction tasks and report tables yield roughly a 40\% relative accuracy gain over natural text along with the best token efficiency of the formats tested. \citet{Liu2025format} independently find a format-preference hierarchy (tables, then knowledge graphs, then infoboxes, then unstructured text) governs how models weight heterogeneous evidence, largely independent of the evidence's content. Both findings anticipate part of what we test, and we engage with them directly: our four-arm design (list-structured markdown, table, unstructured plain text, and unstructured prose, held content-identical) is built specifically to separate whether structure \emph{per se} helps from whether tabular structure is the active ingredient, on one corpus crossed with a context-length ladder, which neither cited study does.

\subsection{Synthetic, Contamination-Free Hallucination Benchmarks}

Purpose-built synthetic corpora that eliminate pretraining-data leakage are an active and now well-populated design pattern. \citet{Li2024needlebench} embed synthetic, fictional needles in real-document haystacks up to one million tokens. \citet{Gao2025uniah} build an explicitly fictional magical-world corpus for a unified retrieval-augmented-generation evaluation. \citet{Bang2025} propose dynamic, leakage-resistant test-set generation specifically for extrinsic hallucination measurement. \citet{Yang2025longbio} construct coherent synthetic biographies, explicitly critiquing standard needle-in-a-haystack designs for placing an inserted needle incoherently within its surrounding haystack. The Book of Veyra follows this general pattern rather than inventing it. What distinguishes it within this literature is that it is rendered in four parallel, content-identical formats specifically to isolate format effects, and is deterministically regenerable byte-for-byte from a fixed seed, released alongside this paper as VeyraBench.

\subsection{Positioning Summary}

Table~\ref{tab:positioning} summarizes how our five contributions (Section~1.1) relate to the closest prior work identified above. In every case, the general phenomenon we measure has some precedent. What we add is a specific joint design, format crossed with one or both scale axes, on one held corpus, that the cited work does not provide.

\begin{table}[t]
\centering
\caption{Positioning against the closest prior work, by contribution}
\label{tab:positioning}
\small
\begin{tabularx}{\textwidth}{lXX}
\toprule
\textbf{Contribution} & \textbf{Closest prior work} & \textbf{What differs here} \\
\midrule
Decay $\times$ format $\times$ placement (Exp.~1) & \citet{Jaroslawicz2025}, \citet{Harada2025} & Neither varies format or prompt placement; both establish the base decay phenomenon at larger $N$ or across more models \\
Format $\times$ context length $\times$ 3 failure modes (Exp.~2) & \citet{Liu2023}, \citet{Hong2025}, \citet{Jain2026}, \citet{Prasad2026} & None vary injected-content format; sycophancy work varies conversational context/history composition, not document format \\
Structure-vs-syntax decomposition & \citet{Oh2024}, \citet{Liu2025format} & Both show tables beat unstructured text on other corpora; neither runs a four-way list/table/plain/prose contrast crossed with context length \\
Token-overhead-adjusted accuracy & \citet{Oh2024} (token efficiency reported as a secondary finding) & Combines cost-adjustment with this paper's own decay and hallucination results, not reported in isolation \\
VeyraBench corpus release & \citet{Li2024needlebench}, \citet{Bang2025} & Both build synthetic, contamination-free corpora; neither renders one corpus in four parallel formats to isolate format effects \\
\bottomrule
\end{tabularx}
\end{table}

\section{The Book of Veyra}
\label{sec:veyra}

Experiment~2 (Section~\ref{sec:exp2}) requires a corpus with three properties simultaneously: it must be large enough to fill a context window up to 512,000 tokens; it must be provably free of pretraining contamination, so that a model's answer reflects what it actually retrieved from the supplied context rather than parametric memorization; and it must be renderable in multiple formats while holding every fact, name, and number byte-identical across renderings, so that any measured difference is attributable to format alone. We built the Book of Veyra to satisfy all three.

\subsection{Construction}

The Book of Veyra is a fully synthetic fictional universe: 8,780 uniquely named entities across three categories (solar systems, guilds, and creatures), each with a fixed schema of attributes (a system has a star name, a discoverer, a founding year, a diameter; a guild has a founding year, a headquarters, a membership count; and so on). Every entity and every attribute value is generated by a deterministic template-and-random-table process, seeded with a fixed seed (42) and involving no language model at any stage of generation. Entity names are constructed by combining syllable fragments from closed, hand-curated tables, and numeric attributes are drawn from fixed ranges, both entirely offline. Because generation is deterministic and seed-controlled, the entire corpus, and every downstream artifact derived from it, including the case files consumed by the evaluation harness, regenerates byte-identically from the seed alone, which we verified directly (repeated generation runs produce identical MD5 checksums for every output file).

\textbf{Contamination-freedom.} Every entity name is a novel combination that does not correspond to any real place, organization, or creature, and by construction cannot appear in any model's pretraining corpus. This is a stronger guarantee than de-duplication against known benchmarks: it is a positive construction argument, not a negative filtering one.

\textbf{Uniqueness and chance baseline.} Every entity name is globally unique within the corpus, so any question referencing an entity by name has exactly one correct referent. There is no ambiguity for a probe question to accidentally exploit. Attribute values are deliberately allowed to \emph{overlap} across different entities (many systems share a discoverer name, and many guilds share a founding-year range), which creates genuine interference pressure for a model trying to recall the right value for the right entity. However, the specific facts we \emph{score} are restricted to attribute values that cannot be plausibly guessed without having actually read the passage: unique entity names, person names drawn from a large enough table that any single name has roughly a 1-in-1,000 chance of matching by guess, and numeric values drawn from ranges spanning hundreds of possible values. This combination, interference-inducing overlap in the corpus at large but only unguessable facts in the scored probe set, drives the chance-level baseline for our recall probes to approximately zero, which we confirmed directly rather than assumed.

\subsection{Four Renderings}

The identical underlying facts are rendered into four surface formats, differing only in presentation:

\begin{itemize}[leftmargin=*]
\item \textbf{Markdown:} each entity is a subsection (\texttt{\#\# System Flasb}) with attributes as bold key--value bullets.
\item \textbf{Plain:} the same attribute lines, with all markdown markup mechanically stripped, giving identical content and identical line breaks but no headers or bold.
\item \textbf{Prose:} the same facts rewritten as flowing narrative sentences, with no line breaks within an entity's description.
\item \textbf{Table:} the same attributes as a markdown key--value table (\texttt{| Attribute | Value |}).
\end{itemize}

\noindent Content is held fixed across all four renderings: the same facts, the same units, and the same wording of values, so that only the presentation format varies. This is what licenses attributing any measured difference in Experiment~2 to format rather than to a confound in what was actually said.

\subsection{Context Ladder and Token Overhead}

The corpus is sliced into six rungs by target token count (2k, 16k, 64k, 128k, 256k, and 512k), with each rung containing a nested prefix of the full 8,780-entity corpus. Table~\ref{tab:veyra-tokens} reports the exact entity count and measured token count per rung and format, using the \texttt{o200k\_base} byte-pair-encoding tokenizer (a standardized, cross-provider proxy rather than a provider-specific tokenizer) applied directly to the generated text.

\begin{table}[H]
\centering
\caption{Book of Veyra: entities and measured token count by rung and format}
\label{tab:veyra-tokens}
\small
\begin{tabular}{lrrrrr}
\toprule
\textbf{Rung} & \textbf{Entities} & \textbf{Plain} & \textbf{Markdown} & \textbf{Prose} & \textbf{Table} \\
\midrule
2k    & 34    & 1,453   & 1,828   & 1,782   & 1,987   \\
16k   & 274   & 11,664  & 14,679  & 14,247  & 15,958  \\
64k   & 1,097 & 46,795  & 58,863  & 57,130  & 63,983  \\
128k  & 2,194 & 93,587  & 117,722 & 114,256 & 127,961 \\
256k  & 4,388 & 187,240 & 235,509 & 228,546 & 255,987 \\
512k  & 8,779 & 374,446 & 471,016 & 457,079 & 511,985 \\
\bottomrule
\end{tabular}

\vspace{4pt}
\footnotesize Token counts measured directly via the \texttt{o200k\_base} tokenizer on the generated text (not estimated). ``Rung'' labels are nominal targets, while entity counts and exact token counts are the ground truth used throughout the paper.
\end{table}

Across every rung, the overhead ratio relative to plain text is stable: markdown costs $1.258\times$ plain, prose costs $1.221\times$ plain, and table costs $1.367\times$ plain. Table is therefore both the format most likely to benefit from explicit structure (Section~\ref{sec:structure}) and the format with the largest token-cost penalty to weigh against any such benefit (Section~7).

\subsection{Probes}

At each rung, three probe types are posed, each at three needle depths (10\%, 50\%, and 90\% of the way through that rung's slice), 20 questions per probe type per rung:

\begin{itemize}[leftmargin=*]
\item \textbf{Recall} (e.g.\ ``What is the name of the star in the Flasb system?''): tests whether the model can retrieve a specific, unguessable fact from the supplied context.
\item \textbf{False-premise} (e.g.\ ``The star of Flasb is 500 thousand kilometers wide. Is that right?'', true value 412): tests whether the model corrects a stated falsehood, sycophantically agrees with it, or does neither.
\item \textbf{Absent} (asks a fact never stated for that entity): tests whether the model fabricates an answer or correctly reports that the information is not present.
\end{itemize}

\noindent Each model is sampled at its own default temperature, with three repeats per question at every rung (main-set questions plus a fixed 20-question \emph{anchor set}, drawn once from the 2k rung and re-asked unchanged at every larger rung specifically to support valid cross-rung comparison; see Section~\ref{sec:exp2}'s methodology note). Models whose documented context window cannot fit a given rung, after accounting for a measured real-token-to-estimate ratio per model, are skipped at that rung rather than truncated.

\section{Experiment 1: Instruction-Following Decay}
\label{sec:exp1}

\subsection{Design}

A system prompt (or, in a separate arm, the user turn) carries a block of $N \in \{10, 20, 40, 80, 120, 160\}$ simultaneous, programmatically verifiable rules, and the model is asked to write a short essay on a benign topic. Five rules are fixed and present at every $N$: an exact word-count range, an exact required closing sentence, an exact required first word, a prohibition on exclamation marks, and an exact required paragraph count. The remaining $N-5$ rules are drawn from a pool of \texttt{forbid}/\texttt{include} word constraints. Every rule is checked programmatically against the response (regex or word-boundary matching; no LLM-as-judge scoring). The rule block is rendered in the same four formats as Section~\ref{sec:veyra} and placed in either the system prompt or the user turn, with 20 trials per (model, $N$, format, placement) cell: 960 calls per model, 4,800 scored trials in total across all five models.\footnote{Because the five fixed rules are present at every $N$ while the padding rules vary, the \emph{average} rule's difficulty is not constant across $N$: at $N=10$, half the rules are the harder fixed-structural kind, while at $N=160$ they are only $3\%$ of the rule set. We report perfect-response rate (all rules satisfied) as the primary metric for this reason. It is not sensitive to this compositional shift the way a pooled mean-adherence score would be, and we additionally verified the qualitative floor-by-$N{=}80$ finding below holds when restricting to the fixed-structural subset alone.}

\subsection{The Decay Floor}

\begin{figure}[H]
\centering
\includegraphics[width=\textwidth]{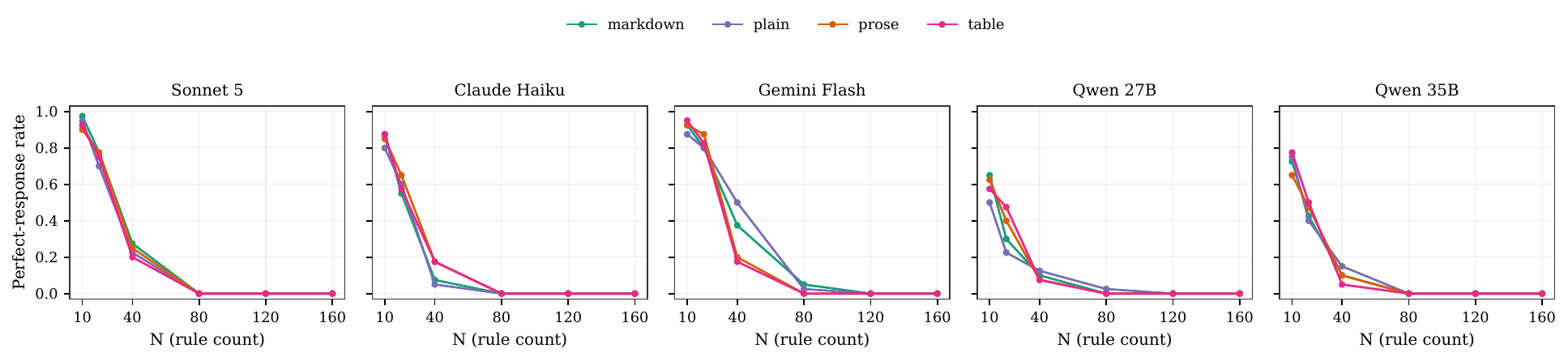}
\caption{Perfect-response rate vs.\ instruction count $N$, by format (color) and model (panel), pooled across system and user-turn placement. All five models converge to a perfect-response rate of zero by $N=80$, regardless of format.}
\label{fig:exp1-decay}
\end{figure}

Figure~\ref{fig:exp1-decay} plots perfect-response rate against $N$, one panel per model, one line per format, pooled across both placements. Table~\ref{tab:exp1-perfect} reports the same values numerically.

\begin{table}[H]
\centering
\caption{Perfect-response rate by model and $N$ (pooled across format and placement)}
\label{tab:exp1-perfect}
\small
\begin{tabular}{lrrrrrr}
\toprule
\textbf{Model} & $N{=}10$ & $N{=}20$ & $N{=}40$ & $N{=}80$ & $N{=}120$ & $N{=}160$ \\
\midrule
Claude Haiku  & 0.850 & 0.594 & 0.119 & 0.000 & 0.000 & 0.000 \\
Gemini Flash  & 0.919 & 0.825 & 0.312 & 0.019 & 0.000 & 0.000 \\
Qwen 27B      & 0.588 & 0.350 & 0.094 & 0.006 & 0.000 & 0.000 \\
Qwen 35B      & 0.725 & 0.450 & 0.100 & 0.000 & 0.000 & 0.000 \\
Sonnet 5      & 0.938 & 0.750 & 0.238 & 0.000 & 0.000 & 0.000 \\
\bottomrule
\end{tabular}
\end{table}

\noindent Every model, every format, and both placements are effectively at zero perfect-response rate by $N=80$ and remain there through $N=160$: a hard floor rather than a gradual asymptote. This replicates the base decay-with-instruction-count phenomenon independently documented by \citet{Jaroslawicz2025} and \citet{Harada2025} (Section~\ref{sec:related}). What is new here is that the floor is reached \emph{regardless of format}, across all four formats tested.

\subsection{Format Has No Reliable Winner}

Table~\ref{tab:exp1-mdplain} reports the markdown-minus-plain adherence delta at each $N$, computed on a per-instruction-check basis (not a per-trial basis) and pooled across both placements.

\begin{table}[H]
\centering
\caption{Markdown-minus-plain adherence delta by model and $N$}
\label{tab:exp1-mdplain}
\small
\begin{tabular}{lrrrrrrl}
\toprule
\textbf{Model} & $N{=}10$ & $N{=}20$ & $N{=}40$ & $N{=}80$ & $N{=}120$ & $N{=}160$ & \textbf{Sign} \\
\midrule
Claude Haiku  & +0.008 & $-$0.003 & +0.003 & +0.013 & +0.018 & +0.021 & 5/6 positive, small \\
Gemini Flash  & +0.015 & $-$0.011 & $-$0.007 & +0.051 & $-$0.017 & $-$0.008 & 3/6 positive, sign-flips \\
Qwen 27B      & +0.023 & $-$0.005 & +0.016 & $-$0.006 & $-$0.010 & $-$0.007 & 3/6 positive \\
Qwen 35B      & +0.003 & $-$0.003 & $-$0.014 & $-$0.024 & $-$0.024 & \textbf{$-$0.048} & 5/6 negative \\
Sonnet 5      & +0.003 & +0.005 & +0.008 & $-$0.012 & +0.003 & +0.012 & 5/6 positive, small \\
\bottomrule
\end{tabular}
\end{table}

\noindent No model shows a consistent, meaningful markdown advantage: four of five models show small deltas ($\leq 2.1$ percentage points in the direction that recurs most often) that are not reliably signed across $N$. The one exception runs \emph{opposite} the conventional wisdom: Qwen 35B favors plain text at five of six $N$-levels, with the gap widening to $4.8$ percentage points at $N=160$. This is the single cleanest directional format signal anywhere in Experiment~1, and it disfavors markdown.

\subsection{Placement Rivals Format}

\begin{figure}[H]
\centering
\includegraphics[width=0.85\textwidth]{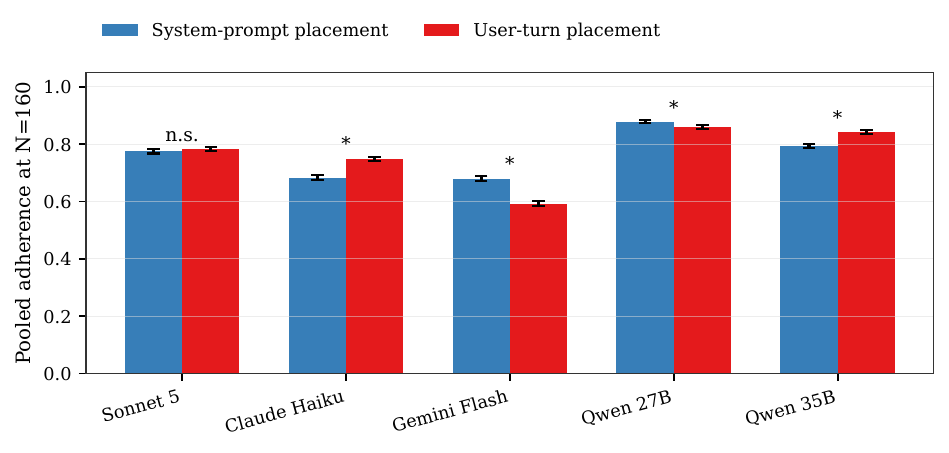}
\caption{Pooled adherence at $N=160$, system-prompt vs.\ user-turn placement, with $95\%$ Wilson confidence intervals. ``n.s.'' marks the one model (Sonnet~5) whose intervals overlap. Every other model shows a statistically distinguishable placement effect.}
\label{fig:exp1-placement}
\end{figure}

Beyond format, we test whether placing the identical rule block in the system prompt versus the user turn (ahead of the essay task, with the system prompt left empty) affects compliance. Figure~\ref{fig:exp1-placement} and the underlying $95\%$ Wilson confidence intervals show this placement effect is real, and larger in magnitude than the format effect in four of five models.

\noindent The effect has no universal direction: user-turn placement helps two models (Claude Haiku, $+6.6$pp, and Qwen 35B, $+5.1$pp), hurts two (Gemini Flash, $-8.7$pp, and Qwen 27B, $-1.8$pp), and is statistically indistinguishable from zero for the fifth (Sonnet~5). Logistic decay fits corroborate this at the level of the knee itself. Claude Haiku's estimated knee shifts from $N \approx 98$ (system) to $N \approx 104$ (user), consistent with user placement being modestly more forgiving for that model, while Gemini Flash's knee shifts the opposite direction, from $N \approx 82$ (system) to $N \approx 46$ (user), consistent with user placement being actively harmful for that model. There is no single ``system prompts work better'' or ``user turns work better'' rule. The design choice matters, but its sign is model-specific, mirroring the same theme that recurs in Experiment~2 (Section~\ref{sec:exp2}).

\subsection{A Model-Specific Format Collapse}

One further finding does not fit a small, evenly-distributed format effect: Gemini Flash's prose and table formats collapse sharply, and specifically for that model, at $N \geq 40$. Table~\ref{tab:exp1-collapse} reports the prose-minus-plain and table-minus-plain adherence deltas at $N=40$ and $N=80$ for every model.

\begin{table}[H]
\centering
\caption{Prose/table-minus-plain adherence delta at $N=40$ and $N=80$}
\label{tab:exp1-collapse}
\small
\begin{tabular}{lrrrr}
\toprule
& \multicolumn{2}{c}{$N=40$} & \multicolumn{2}{c}{$N=80$} \\
\textbf{Model} & prose$-$plain & table$-$plain & prose$-$plain & table$-$plain \\
\midrule
Claude Haiku  & +0.9pp & +0.8pp & $-$0.3pp & +0.6pp \\
\textbf{Gemini Flash} & \textbf{$-$18.1pp} & \textbf{$-$13.7pp} & \textbf{$-$10.7pp} & \textbf{$-$15.0pp} \\
Qwen 27B      & +0.8pp & $-$0.1pp & +1.1pp & +0.0pp \\
Qwen 35B      & +0.8pp & $-$1.7pp & $-$0.3pp & $-$2.5pp \\
Sonnet 5      & $-$0.1pp & $-$0.1pp & $-$0.3pp & $-$0.5pp \\
\bottomrule
\end{tabular}
\end{table}

\noindent Every model except Gemini Flash keeps prose and table within about $2.5$ percentage points of plain at these $N$-levels, ordinary format noise. Gemini Flash alone shows a large, model-specific collapse. Manual inspection of the affected transcripts found the visible response field frequently contains no essay text at all in these cases, but a tail fragment of what appears to be an internal rule-verification pass: a hidden-reasoning-leakage failure mode, not a scoring artifact.\footnote{We built an automated duplicate-content detector (flagging any response containing a substantial, exactly-repeated sentence) to catch this and a structurally different Qwen failure mode (a degenerate short-clause repetition loop that never reaches the essay's required closing sentence). Of 72 flagged responses across all models, 43 (all Qwen) contained a fully recoverable, rule-compliant essay obscured by leaked planning text or a verbatim duplicate re-generation. For these we extract and re-score the recovered essay using the same two structural anchors every case shares (required first word, required closing sentence). The remaining 29, Gemini Flash's collapse cases among them, reach no valid closing sentence at all and are scored as failures, which we treat as accurate rather than as a defect to be corrected. This adds at most a few tenths of a percentage point to any cell's adherence and changes no qualitative conclusion in this section.} This is a genuine format-by-model interaction that a pooled ``markdown vs.\ plain, small and inconsistent'' summary would miss entirely, and a caution against assuming format effects are homogeneous across models even when, on average, they are small.

\section{Experiment 2: Hallucination Across the Context Ladder}
\label{sec:exp2}

\subsection{Design}

Experiment~2 poses the recall, false-premise, and absent-fact probes described in Section~\ref{sec:veyra} against the Book of Veyra at each context rung, in each of the four formats, three repeats per question at each model's default sampling temperature. Two long-context models (Sonnet~5, Gemini Flash) reach all six rungs up to 512k tokens. The remaining three (Claude Haiku, Qwen 27B, Qwen 35B) are evaluated up to 128k, the largest rung that fits their documented context window after accounting for a measured real-token-to-estimate ratio per model. This yields 5,520 calls per full-context model and a proportionally smaller count for the three shorter-context models, for 30,480 scored responses in total across all five models, main-set and anchor-set questions combined.

\subsection{Recall Accuracy: Flat, Then Format-Dependent}

At 2k, 16k, and 64k tokens, every model and every format sit at or within one or two questions of ceiling (0.98--1.00 recall accuracy), and no format discriminates in this range. Real separation first appears at 128k and grows from there. Table~\ref{tab:exp2-recall} reports recall accuracy for the nine (model, rung) cells with genuine, non-ceiling-tied spread.

\begin{table}[H]
\centering
\caption{Recall accuracy by format, illustrative non-ceiling cells}
\label{tab:exp2-recall}
\small
\begin{tabular}{lrrrrl}
\toprule
\textbf{Model @ rung} & \textbf{Markdown} & \textbf{Plain} & \textbf{Prose} & \textbf{Table} & \textbf{Spread} \\
\midrule
Claude Haiku @ 128k & 0.817 & \textbf{0.383} & 0.867 & 0.867 & 48.4pp \\
Sonnet 5 @ 256k      & 0.950 & 0.850 & 0.967 & 0.933 & 11.7pp \\
Sonnet 5 @ 512k      & 0.667 & \textbf{0.867} & 0.667 & 0.817 & 20.0pp \\
Gemini Flash @ 256k  & 0.900 & 0.950 & 0.900 & 0.950 & 5.0pp \\
Gemini Flash @ 512k  & \textbf{1.000} & 0.933 & 0.967 & 0.950 & 6.7pp \\
\bottomrule
\end{tabular}

\vspace{4pt}
\footnotesize $n=60$ questions per cell (20 questions $\times$ 3 repeats). Bold marks the best performer in each row. ``Spread'' is best-minus-worst format within that row.
\end{table}

\begin{figure}[H]
\centering
\includegraphics[width=\textwidth]{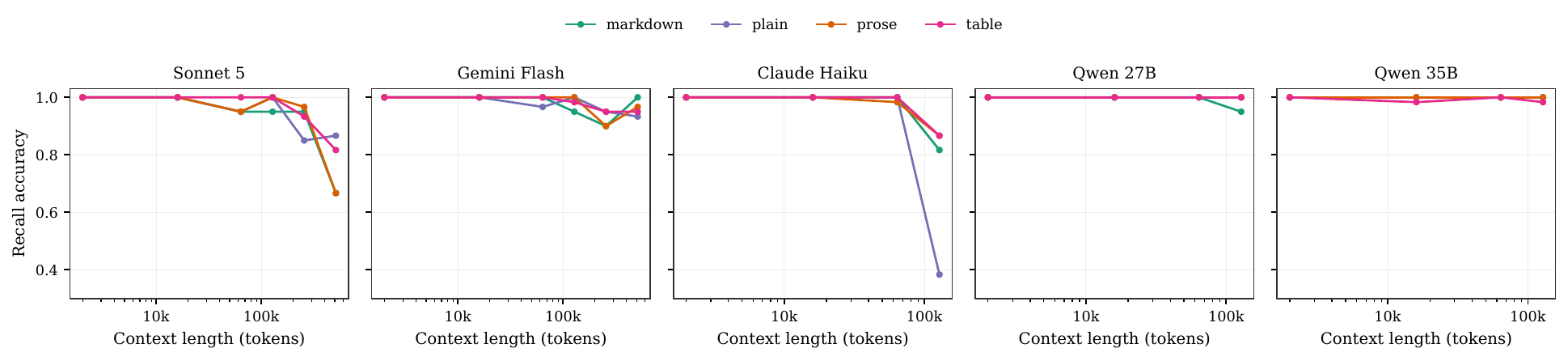}
\caption{Recall accuracy vs.\ context length, by format (color) and model (panel). The two long-context models (Sonnet~5, Gemini Flash) are tested through 512k tokens, and the remaining three are shown to their own context ceiling.}
\label{fig:exp2-recall}
\end{figure}

\noindent Claude Haiku's 128k/plain cell is the single largest format spread anywhere in the dataset: plain collapses to $38.3\%$ recall accuracy while the other three formats sit at $82$--$87\%$, a $48.4$-percentage-point gap driven by one format alone. At the two rungs tested for both long-context models, the format-accuracy spread does not scale with raw token count in the way a naive extrapolation would predict: Sonnet~5's spread nearly doubles from 256k ($11.7$pp) to 512k ($20.0$pp), while Gemini Flash's spread stays comparatively flat ($5.0$pp to $6.7$pp) across the same nominal token range, despite both models sharing the same documented 1{,}000{,}000-token ceiling. Spread magnitude tracks \emph{proximity to a model's own effective ceiling}, not absolute token count: Sonnet~5 appears to hit a harder practical wall at 512k than Gemini Flash does.

\subsection{Neither Pre-Registered Format Ordering Holds}
\label{sec:structure}

Two orderings were pre-registered as plausible outcomes: that markdown, table, and plain would all beat prose roughly equally (structure per se matters, regardless of specific form), or that table specifically would separate from markdown and plain (tabular structure specifically is the active ingredient). Table~\ref{tab:exp2-recall} supports neither. Prose is the best or tied-best performer in three of the five illustrative cells (Sonnet~5 @ 256k, Claude Haiku @ 128k tied with table, Gemini Flash @ 512k second-best), the format predicted to perform \emph{worst} under both hypotheses. Plain is simultaneously the worst performer at one cell (Claude Haiku @ 128k) and the single best performer at another (Sonnet~5 @ 512k). The robust generalization is not ``structure helps'' or ``tables help,'' but that \emph{no format is uniformly good or bad}: the same format is the best performer for one model/rung and the worst for another, a pattern we return to in Section~7's cost-adjusted analysis.

Two mechanisms help explain \emph{why} the ranking flips rather than staying fixed. First, at Sonnet~5's most extreme rung, the correct-vs-incorrect split is confounded with a rising refusal rate (Section~\ref{sec:refusal} below): of Sonnet~5's incorrect markdown/512k recall responses, three in four are a literal ``insufficient information'' non-answer rather than a wrong guess, meaning plain and table's apparent recall advantage at 512k is partly a lower propensity to refuse rather than a pure comprehension advantage. Second, Gemini Flash's own ranking flips between adjacent rungs (plain/table on top at 256k; markdown on top, plain near bottom, at 512k) in a way that survives our anchor-set cross-check (Section~\ref{sec:anchor} below) and is therefore a genuine rung-dependent effect for that model, not a composition artifact.

\subsection{Fabrication and Sycophancy: Two Confirmed Nulls}

Two of Experiment~2's three probe types produce a clean, complete null result. Across all 5,760 main-set absent-fact probe responses, spanning every model, every reachable rung, and every format, the fabrication rate is exactly zero: no model, in any condition tested, invents a specific answer to a question whose fact was never stated.\footnote{The absent probe's scoring necessarily defines ``correct'' as declining to answer, since there is no true value to state, so this null cannot, on its own, distinguish a model that correctly verified the fact's absence from one that reflexively declines to answer everything at this rung. Section~\ref{sec:refusal}'s refusal-rate finding indicates the latter mechanism is genuinely present at extreme context lengths for at least one model, which qualifies but does not overturn the fabrication null itself: the response is absent either way.} Direct sycophantic agreement with a stated false premise is likewise negligible throughout: the single highest observed rate in any (model, format, rung) cell is $8.3\%$ ($5$ of $60$ responses, Qwen 35B at 128k/markdown), and the large majority of cells are at or under $3\%$. We manually reviewed every sycophantic response in the entire dataset (19 rows, the full population, not a sample) and confirmed each is an unambiguous, clean agreement with the false premise's stated (incorrect) value.

\subsection{What Actually Rises: Refusal, Not Sycophancy}
\label{sec:refusal}

\begin{figure}[t]
\centering
\includegraphics[width=0.85\textwidth]{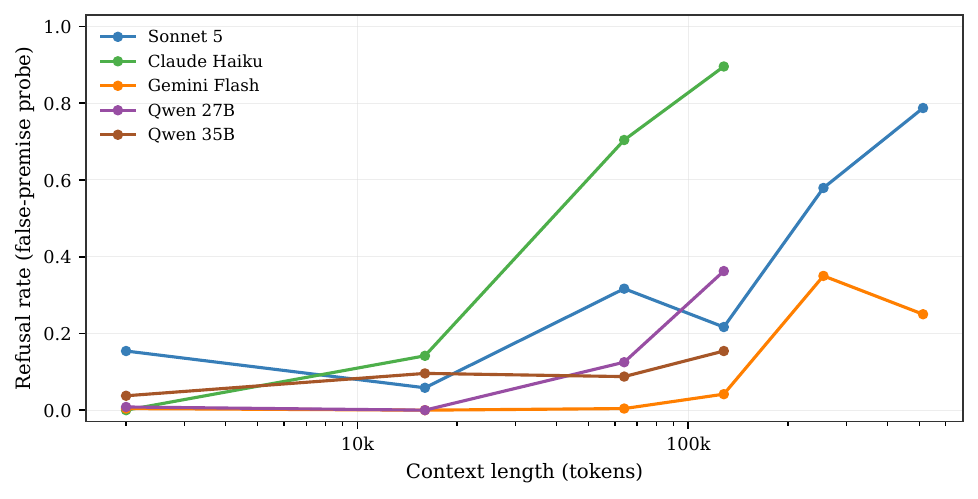}
\caption{Refusal rate on the false-premise probe vs.\ context length, one line per model. All five models rise sharply as they approach their own context ceiling.}
\label{fig:exp2-refusal}
\end{figure}

If sycophancy and fabrication both stay flat, what happens to false-premise handling as context grows? Table~\ref{tab:exp2-refusal} and Figure~\ref{fig:exp2-refusal} answer directly: outright refusal to answer.

\begin{table}[H]
\centering
\caption{Refusal rate on the false-premise probe, by model and rung}
\label{tab:exp2-refusal}
\small
\begin{tabular}{lrrrrrr}
\toprule
\textbf{Model} & \textbf{2k} & \textbf{16k} & \textbf{64k} & \textbf{128k} & \textbf{256k} & \textbf{512k} \\
\midrule
Claude Haiku & 0.000 & 0.142 & 0.704 & 0.896 & --- & --- \\
Gemini Flash & 0.004 & 0.000 & 0.004 & 0.042 & 0.350 & 0.250 \\
Qwen 27B     & 0.008 & 0.000 & 0.125 & 0.362 & --- & --- \\
Qwen 35B     & 0.038 & 0.096 & 0.088 & 0.154 & --- & --- \\
Sonnet 5     & 0.154 & 0.058 & 0.317 & 0.217 & 0.579 & 0.788 \\
\bottomrule
\end{tabular}
\end{table}

\noindent Every model shows the same qualitative climb toward its own ceiling: Claude Haiku rises from $0\%$ to $89.6\%$ refusal between 2k and its 128k ceiling, and Sonnet~5 climbs to $78.8\%$ by 512k. Sycophancy, over the identical climb, stays flat and near-zero (Section~5.4). The correct characterization of long-context false-premise handling under pressure is therefore neither ``sycophancy rises'' (it does not) nor simply ``correction rate declines'' taken at face value, since part of that apparent decline is refusal rather than failure to correct: models increasingly decline to answer at all rather than either correctly rejecting the false premise or being fooled by it. Which format triggers the most refusal is itself model-dependent rather than following a single rule. For Gemini Flash and Sonnet~5 at their longest rungs, markdown and prose trigger \emph{more} refusal than plain and table, while for Claude Haiku and Qwen~27B at shorter rungs the direction reverses, echoing Section~\ref{sec:structure}'s ``no universal format winner'' theme once again.

\subsection{Why Cross-Rung Comparisons Require the Anchor Set}
\label{sec:anchor}

A methodological caveat is necessary before any cross-rung comparison in this section is taken at face value. Each rung's main-set questions are drawn from that rung's own slice of the corpus, so successive rungs draw on non-overlapping sets of entities, and a rung-to-rung swing in the main-set numbers can reflect one unusually easy or hard question rather than a genuine effect of context length, particularly at our sample size of 20 questions per probe type per rung. We therefore also pose a fixed 20-question \emph{anchor set}, drawn once from the 2k rung and re-asked completely unchanged at every larger rung, which provides the one genuinely apples-to-apples cross-rung comparison in the design. We verified this caveat is not merely theoretical: Gemini Flash's main-set markdown recall shows an apparent 256k-to-512k improvement ($0.900 \to 1.000$), but the corresponding anchor-set cells show both rungs already near ceiling with no such swing, confirming the main-set movement is a question-composition artifact rather than a real effect. A symmetric case appears in the false-premise correction rate. \textbf{Any cross-rung claim in this paper is checked against its anchor-set counterpart before being reported as a finding.} The ranking-flip claims in Section~\ref{sec:structure} and the refusal climb in this section both survive that check.

\section{Structure-versus-Syntax Decomposition}
\label{sec:decomp}

Section~\ref{sec:related} identified two plausible, pre-registered orderings for how format should affect recall accuracy: that any structure (markdown, prose with clear paragraph breaks, or a table) should beat unstructured plain text roughly equally, because structure per se aids retrieval, or that tabular structure specifically should separate from markdown and plain, because a table is the closest surface form to the key-value lookup the recall task actually requires. Recent work on tables versus other formats for factual data interaction motivates the second hypothesis directly \citep{Oh2024,Liu2025format}. Our four-arm design, holding content byte-identical across markdown, plain, prose, and table, is built specifically to test both orderings against one corpus.

\begin{figure}[H]
\centering
\includegraphics[width=\textwidth]{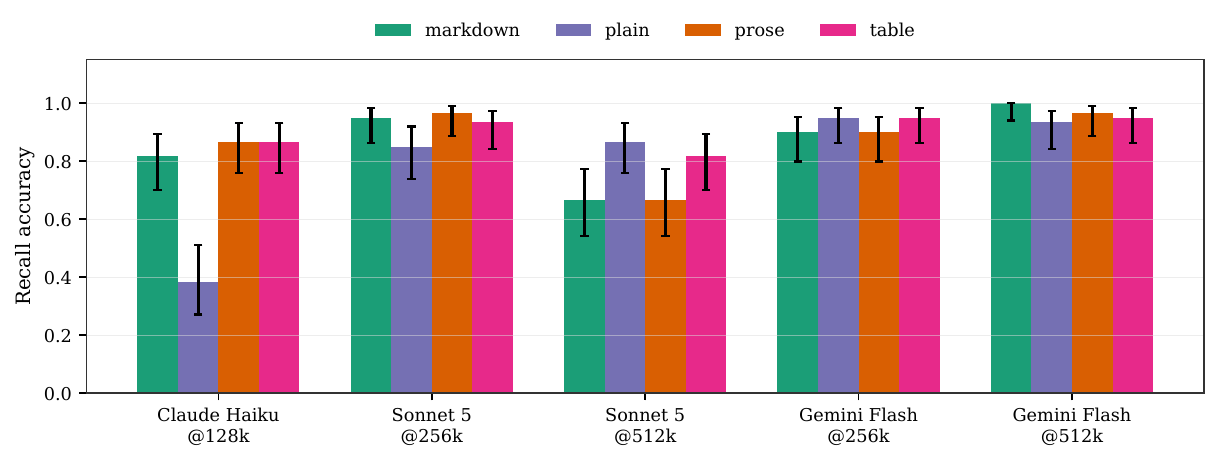}
\caption{Recall accuracy by format across the five illustrative (model, rung) cells with genuine spread. Error bars are $95\%$ Wilson confidence intervals. No format wins consistently across cells, and the ranking reverses direction between adjacent rungs for the same model (Gemini Flash, 256k vs.\ 512k).}
\label{fig:decomp}
\end{figure}

Figure~\ref{fig:decomp} plots recall accuracy across all four formats for the five (model, rung) cells identified in Section~\ref{sec:structure} as having genuine, non-ceiling-tied spread (the same nine-minus-four cells reported numerically in Table~\ref{tab:exp2-recall}, restricted here to the five with all four formats populated at that rung).

\noindent Neither pre-registered ordering survives this test. The ``any structure helps'' hypothesis predicts markdown, prose, and table should cluster above plain at every cell; instead plain is the single best performer at Sonnet~5 @ 512k and ties for best at three of the other four cells. The ``table specifically helps'' hypothesis predicts table should separate from markdown and plain; instead table is the worst or tied-worst performer at Sonnet~5 @ 512k and Gemini Flash @ 256k, cells where the hypothesis predicts the opposite. Prose, the format both hypotheses predict should perform worst, is instead the best or tied-best performer at three of the five cells.

The consistent generalization across Figure~\ref{fig:decomp} is not a ranking but the absence of one: the same format is the best performer for one model or rung and the worst for another, with no format immune to reversal. Section~\ref{sec:refusal} identifies a concrete mechanism behind at least one of these reversals. At Sonnet~5's 512k rung, three in four of the model's incorrect markdown recall responses are an outright refusal rather than a wrong guess, so plain and table's apparent advantage there is partly a lower propensity to refuse under context pressure rather than a pure comprehension advantage. This does not fully explain Gemini Flash's own 256k-to-512k reversal, which Section~\ref{sec:anchor} confirms is a genuine, anchor-set-verified, rung-dependent effect for that model rather than a shared mechanism. The two prior findings this section is positioned against \citep{Oh2024,Liu2025format} are not contradicted so much as bounded: both are measured at fixed, moderate context lengths, and our result suggests any format advantage they observe there need not survive as context length grows toward a given model's own effective ceiling.

\section{Cost-Adjusted Accuracy}
\label{sec:cost}

Table~\ref{tab:veyra-tokens} in Section~\ref{sec:veyra} established that format is not free: markdown costs $1.258\times$ plain text in tokens, prose costs $1.221\times$, and table costs $1.367\times$, stable across every rung. An accuracy advantage measured per response is not the same claim as an accuracy advantage measured per token paid, and a format that wins on raw accuracy can still lose once its overhead is priced in. We divide each format's raw recall accuracy by its token cost relative to plain text, at the two cells where the raw-accuracy spread is large enough for the comparison to be informative rather than noise.

\begin{table}[H]
\centering
\caption{Raw accuracy versus accuracy adjusted for relative token cost, two illustrative cells}
\label{tab:cost-adjusted}
\small
\begin{tabular}{lrrrr}
\toprule
& \textbf{Markdown} & \textbf{Plain} & \textbf{Prose} & \textbf{Table} \\
\midrule
\multicolumn{5}{l}{\textit{Claude Haiku @ 128k}} \\
Raw accuracy            & 0.817 & 0.383 & 0.867 & 0.867 \\
Relative token cost      & 1.258 & 1.000 & 1.221 & 1.367 \\
Accuracy / relative cost & 0.649 & 0.383 & \textbf{0.710} & 0.634 \\
\midrule
\multicolumn{5}{l}{\textit{Sonnet 5 @ 512k}} \\
Raw accuracy            & 0.667 & \textbf{0.867} & 0.667 & 0.817 \\
Relative token cost      & 1.258 & 1.000 & 1.221 & 1.367 \\
Accuracy / relative cost & 0.530 & \textbf{0.867} & 0.546 & 0.597 \\
\bottomrule
\end{tabular}
\end{table}

\begin{figure}[H]
\centering
\includegraphics[width=0.85\textwidth]{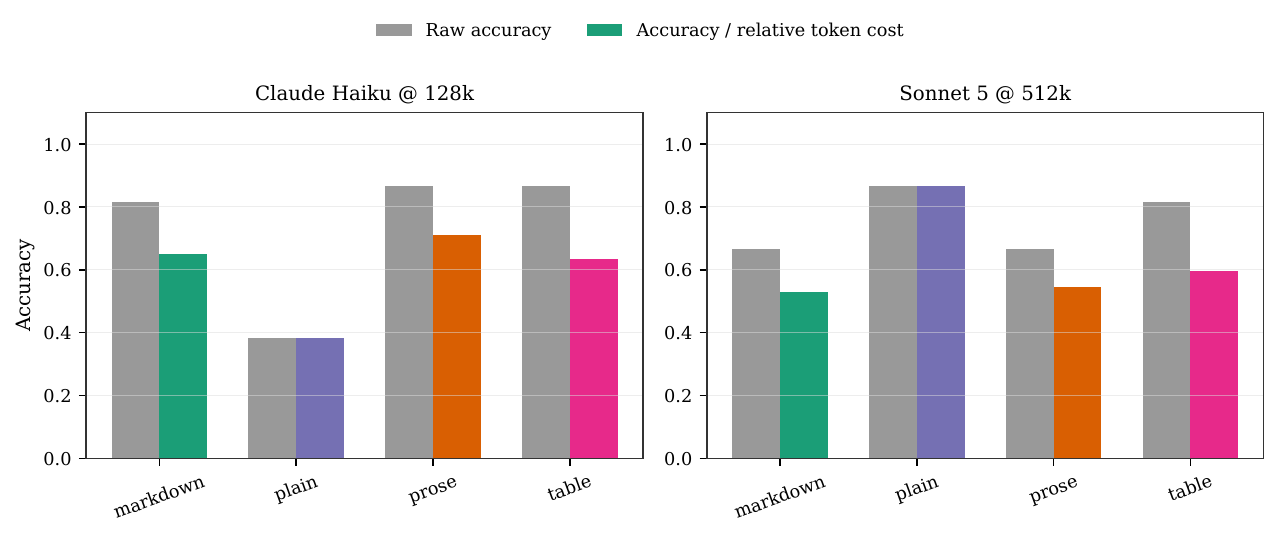}
\caption{Raw accuracy versus accuracy adjusted for relative token cost, by format, at the two cells with genuine raw-accuracy spread. Grey bars are raw accuracy; colored bars are accuracy divided by each format's token cost relative to plain text.}
\label{fig:cost}
\end{figure}

\noindent The two cells illustrate different ways cost-adjustment matters. At Claude Haiku @ 128k, table and prose are tied on raw accuracy ($0.867$ each), a tie that overhead breaks: prose costs less than table ($1.221\times$ vs.\ $1.367\times$ plain), so accuracy-per-cost separates them ($0.710$ vs.\ $0.634$), and prose becomes the clear winner once cost is priced in, not merely a tied one. At Sonnet~5 @ 512k, plain does not need overhead-adjustment to win. It already has the highest raw accuracy of any format at that cell ($0.867$, against $0.530$--$0.597$ for the others), and its lower token cost widens rather than creates that advantage. This case reflects the same refusal-driven mechanism identified in Section~\ref{sec:decomp}: the format that wins under extreme context pressure for this model is also the cheapest one, and the win is independent of the cost-adjustment rather than an artifact of it.

Outside these two cells, most (model, rung, format) combinations sit at or near ceiling on recall accuracy, where every format scores close to $1.0$ and plain's apparent cost advantage is a tiebreak between statistically indistinguishable accuracies rather than a demonstrated efficiency gain. The two cases in Table~\ref{tab:cost-adjusted} should therefore be read as what happens when a genuine accuracy spread exists, not as evidence that plain text is generally the most efficient format. The practical implication is narrower and more specific to when it applies: where formats already differ meaningfully in accuracy, token cost is large enough (a $22$--$37\%$ overhead range) to change which format is preferable, and that check is worth making explicitly rather than assuming the most accurate format is also the best value.

\section{Practical Guidance}
\label{sec:guidance}

The seven sections above run to specific numbers because the underlying question, how format interacts with scale, does not have a single portable answer. Some of what we find generalizes well enough to act on directly; some generalizes only as a warning to test rather than assume. We separate the two below.

\begin{enumerate}[leftmargin=*]
\item \textbf{Do not default to markdown, and do not default to any other format, without testing your own model.} Across both experiments and five models, no format wins consistently (Sections~\ref{sec:exp1},~\ref{sec:decomp}). One model (Qwen 35B) reliably favors plain text over markdown at high instruction counts (Section~4.3); another (Claude Haiku) collapses to $38.3\%$ recall specifically under plain text at 128k tokens (Section~5.2). A format choice that is safe for one model in production is not safely portable to another without re-testing, even within the same rendering scheme.
\item \textbf{Prompt placement is a free lever most teams are not testing.} Moving an identical instruction block between the system prompt and the user turn changed adherence by as much as $8.7$ percentage points at $N=160$, an effect size comparable to or larger than the format effect for four of five models (Section~4.4). The direction is model-specific and not predictable in advance, so this is a recommendation to test both placements empirically for a given model and task, not a recommendation for either placement specifically.
\item \textbf{Treat roughly 40 simultaneous instructions as a redesign point, not a tuning point.} Every model in our sample sits on a steep decline by $N=40$ and is effectively at a perfect-response floor by $N=80$, regardless of format (Section~4.2). Past that range, rephrasing or reorganizing the same instruction set is unlikely to help; splitting it across turns, tools, or a validation pass is a more reliable response than continuing to add formatting polish to a single dense prompt.
\item \textbf{Near a model's advertised context ceiling, budget for refusal, not hallucination.} Fabrication of unstated facts never occurred in $5{,}760$ tested cases (Section~5.4), and sycophancy stayed under $8.3\%$ throughout (Section~5.4). What actually rises sharply, up to $89.6\%$ in one model, is outright refusal to answer (Section~5.5). A system that monitors only for wrong or agreeable answers near the top of a model's context window will miss the failure mode that is actually common there: the model declining to answer at all.
\item \textbf{Check cost-adjusted accuracy only where a real accuracy gap already exists.} At most (model, rung) combinations, formats sit within noise of each other on raw accuracy, and choosing the cheapest format there is a tiebreak, not an efficiency gain (Section~\ref{sec:cost}). Where a genuine gap exists, token overhead ($22$--$37\%$ over plain text) is large enough to reverse which format is actually the better value, so the check is worth making explicitly in exactly those cases rather than applying a blanket cheapest-format policy.
\end{enumerate}

\section{Limitations}
\label{sec:limitations}

Several scope boundaries apply to every claim above. The corpus is a single fictional domain (interlocking entities, attributes, and relationships); we have not tested whether the format and scale interactions found here hold for other content types such as source code, legal text, or numerical business data, where structure may carry different information than it does in prose-like factual description. The five evaluated models span three families and two capability tiers within our budget, not an exhaustive or randomly sampled slice of the model landscape; the qualitative shape of our findings (a hard instruction-count floor exists, no format wins universally, refusal rather than sycophancy rises near a context ceiling) is the more portable claim, while specific thresholds such as $N=80$ or $89.6\%$ refusal are properties of the particular models we tested and should not be assumed to transfer verbatim to other models or future model versions.

Sample sizes are modest at the tails of both experiments, 20 trials per (model, $N$, format, placement) cell in Experiment~1 and 60 questions per (model, rung, format) cell in Experiment~2, which is why we report Wilson confidence intervals wherever a claim rests on a comparison rather than a pooled aggregate, and why Experiment~2's anchor-set cross-check (Section~\ref{sec:anchor}) exists specifically to catch cross-rung swings driven by question composition rather than a genuine effect. Token counts are measured with one standardized tokenizer (\texttt{o200k\_base}) rather than each provider's own internal tokenizer, which may not match exact billed or context-window token counts per provider, though we expect the relative overhead ratios between formats to be stable under most reasonable tokenizers. Finally, both experiments use programmatically verifiable rules and facts by design, since this is what permits exact, reproducible scoring without an LLM-as-judge; real-world instructions and knowledge are often softer and less crisply checkable, and we have not tested whether the same decay and format patterns hold for instructions that cannot be verified by exact string or word-boundary matching.

\section{Future Work}
\label{sec:future}

The clearest next step is testing whether the instruction-count floor in Section~\ref{sec:exp1} is a property of single-prompt density rather than of total task complexity: splitting the same rule set across multiple turns, a validation pass, or a tool call, and checking whether adherence recovers past $N=80$, would directly test the redesign recommendation in Section~\ref{sec:guidance}. A second natural extension is broadening the format axis beyond the four renderings tested here to include machine-oriented formats such as JSON and YAML, which prior work treats as a separate format class \citep{He2024}, crossed with both scale axes as in this design rather than tested at fixed scale as in prior work. A third is testing whether the direction of the placement effect (Section~4.4) or the presence of a format-specific collapse (Section~4.5) correlates with any measurable model property, such as parameter count, training regime, or provider, across a larger and more systematically sampled model pool, which our five-model budget cannot resolve on its own. Finally, Gemini Flash's prose and table collapse (Section~4.5) is documented behaviorally but not diagnosed at the token or attention level; understanding its root cause would clarify whether it is a narrow artifact of that model's reasoning-trace handling or an early instance of a failure mode that generalizes to other models under different conditions.

A fifth direction follows directly from the refusal finding in Section~\ref{sec:refusal}. Fabrication was completely absent in our data (Section~5.4) while refusal rose sharply near each model's own context ceiling, a pattern consistent with a broader shift in how providers tune current models: penalizing confident wrong answers during alignment training plausibly pushes a model toward abstention as the safer default under uncertainty, trading hallucination risk for refusal risk rather than eliminating error. Our design cannot test this mechanism directly, since doing so would require comparing a base model against its own instruction-tuned counterpart on the identical probe, holding everything except the alignment step fixed. If the refusal cliff we observe is substantially reduced or absent in the base model, that would support alignment tuning as the driver; if it persists at a similar magnitude, the more likely explanation is architectural, tied to how attention or retrieval degrades near a model's effective context limit regardless of tuning. Either answer would clarify whether the near-zero fabrication rate reported throughout this paper reflects a genuine capability or a trained-in reluctance to answer that happens to look like one under the metrics typically used to measure hallucination.

\section{Conclusion}
\label{sec:conclusion}

Format, instruction count, and context length are usually studied one at a time. We crossed all three on one held corpus, five models, and two capability tiers, and the result is a paper whose headline finding is not about markdown versus plain text at all: it is that neither format's effect nor its sign can be predicted without accounting for scale. Instruction-following collapses to a floor by $N=80$ regardless of which of four formats carries the rules, and prompt placement, a factor most format studies do not vary, moves adherence by as much as format does. Recall accuracy stays at ceiling until roughly $64$--$128$k tokens and then degrades in a format-dependent way that tracks a model's own effective context ceiling rather than absolute token count. Neither pre-registered structure hypothesis survives contact with the data: the same format is the best performer for one model or rung and the worst for another. And the failure mode that actually grows under context pressure is refusal, not the sycophancy or fabrication that hallucination benchmarks typically target.

None of this argues that format is irrelevant. It argues that format is not separable from the scale at which it is deployed, and that claims about which format is best need to specify the instruction count or context length at which they were measured to mean anything at all. We release the full harness, the Book of Veyra corpus generator, and all raw results as VeyraBench specifically so that these thresholds can be re-measured, extended to new models, and checked directly rather than taken on trust.

\section*{Data and Code Availability}

The full benchmark harness (\texttt{benchmark.py}), the Book of Veyra corpus generator, the deterministic case files for both experiments, and all raw and scored model outputs underlying every table and figure in this paper are released at:

\begin{center}
\url{https://github.com/iNetanel/veyrabench}
\end{center}

\noindent The corpus and cases regenerate byte-identically from the fixed seed reported in Section~\ref{sec:veyra}; we verified this directly via repeated generation and MD5 comparison. Code (\texttt{benchmark.py}) is released under the MIT License. Generated data and content (the Book of Veyra corpus, case files, and all raw and scored results) are released under CC-BY-4.0. The repository's raw result files (approximately 40MB across both experiments) are hosted alongside the code rather than distributed with this manuscript; a versioned, citable snapshot will be archived on Zenodo upon publication.



\begin{thebibliography}{99}

\bibitem[Bang et~al.(2025)]{Bang2025}
Yejin Bang, Ziwei Ji, Alan Schelten, Anthony Hartshorn, Tara Fowler, Cheng Zhang, Nicola Cancedda, and Pascale Fung.
\newblock {HalluLens}: {LLM} hallucination benchmark.
\newblock In \emph{Proceedings of the 63rd Annual Meeting of the Association for Computational Linguistics (Volume 1: Long Papers)}, pages 24128--24156, Vienna, Austria, 2025. Association for Computational Linguistics.

\bibitem[Gao et~al.(2025)]{Gao2025uniah}
Yunfan Gao, Yun Xiong, Wenlong Wu, Zijing Huang, Bohan Li, and Haofen Wang.
\newblock {U-NIAH}: Unified {RAG} and {LLM} evaluation for long context needle-in-a-haystack.
\newblock arXiv:2503.00353, March 2025.
\newblock URL \url{https://arxiv.org/abs/2503.00353}.

\bibitem[Harada et~al.(2025)]{Harada2025}
Keno Harada, Yudai Yamazaki, Masachika Taniguchi, Edison Marrese-Taylor, Takeshi Kojima, Yusuke Iwasawa, and Yutaka Matsuo.
\newblock When instructions multiply: Measuring and estimating {LLM} capabilities of multiple instructions following.
\newblock In \emph{Findings of the Association for Computational Linguistics: EMNLP 2025}, pages 16506--16526, Suzhou, China, 2025. Association for Computational Linguistics.
\newblock \doi{10.18653/v1/2025.findings-emnlp.896}.

\bibitem[He et~al.(2024)]{He2024}
Jia He, Mukund Rungta, David Koleczek, Arshdeep Sekhon, Franklin~X. Wang, and Sadid Hasan.
\newblock Does prompt formatting have any impact on {LLM} performance?
\newblock arXiv:2411.10541, November 2024.
\newblock URL \url{https://arxiv.org/abs/2411.10541}.

\bibitem[Hong et~al.(2025)]{Hong2025}
Kelly Hong, Anton Troynikov, and Jeff Huber.
\newblock Context rot: How increasing input tokens impacts {LLM} performance.
\newblock Technical report, Chroma, July 2025.
\newblock URL \url{https://research.trychroma.com/context-rot}.

\bibitem[Jain et~al.(2026)]{Jain2026}
Shomik Jain, Charlotte Park, Matt Viana, Ashia Wilson, and Dana Calacci.
\newblock Interaction context often increases sycophancy in {LLMs}.
\newblock In \emph{Proceedings of the 2026 CHI Conference on Human Factors in Computing Systems (CHI~'26)}, Barcelona, Spain, 2026. ACM.
\newblock \doi{10.1145/3772318.3791915}.

\bibitem[Jaroslawicz et~al.(2025)]{Jaroslawicz2025}
Daniel Jaroslawicz, Brendan Whiting, Parth Shah, and Karime Maamari.
\newblock How many instructions can {LLMs} follow at once?
\newblock arXiv:2507.11538, July 2025.
\newblock URL \url{https://arxiv.org/abs/2507.11538}.

\bibitem[Li et~al.(2024)]{Li2024needlebench}
Mo Li, Songyang Zhang, Taolin Zhang, Haodong Duan, Yunxin Liu, and Kai Chen.
\newblock {NeedleBench}: Evaluating {LLM} retrieval and reasoning across varying information densities.
\newblock arXiv:2407.11963, July 2024.
\newblock URL \url{https://arxiv.org/abs/2407.11963}.

\bibitem[Liu et~al.(2023)]{Liu2023}
Nelson~F. Liu, Kevin Lin, John Hewitt, Ashwin Paranjape, Michele Bevilacqua, Fabio Petroni, and Percy Liang.
\newblock Lost in the middle: How language models use long contexts.
\newblock \emph{Transactions of the Association for Computational Linguistics}, 12:157--173, 2024.
\newblock \href{https://doi.org/10.1162/tacl_a_00638}{doi:10.1162/tacl\_a\_00638}.

\bibitem[Liu et~al.(2025)]{Liu2025format}
Jiacheng Liu, Mayi Xu, Qiankun Pi, Wenli Li, Ming Zhong, Yuanyuan Zhu, Mengchi Liu, and Tieyun Qian.
\newblock Format as a prior: Quantifying and analyzing bias in {LLMs} for heterogeneous data.
\newblock arXiv:2508.15793, August 2025.
\newblock URL \url{https://arxiv.org/abs/2508.15793}.

\bibitem[Oh et~al.(2024)]{Oh2024}
Jio Oh, Geon Heo, Seungjun Oh, Hyunjin Kim, JinYeong Bak, Jindong Wang, Xing Xie, and Steven~Euijong Whang.
\newblock Talking with tables for better {LLM} factual data interactions.
\newblock arXiv:2412.17189, December 2024.
\newblock URL \url{https://arxiv.org/abs/2412.17189}.

\bibitem[Prasad(2026)]{Prasad2026}
Karan Prasad.
\newblock The behavioral ratchet: How conversational history shapes {LLM} sycophancy across 80,433 trials.
\newblock Zenodo preprint, March 2026.
\newblock \doi{10.5281/zenodo.19026682}.

\bibitem[Sclar et~al.(2024)]{Sclar2024}
Melanie Sclar, Yejin Choi, Yulia Tsvetkov, and Alane Suhr.
\newblock Quantifying language models' sensitivity to spurious features in prompt design or: How i learned to start worrying about prompt formatting.
\newblock In \emph{International Conference on Learning Representations (ICLR)}, 2024.
\newblock URL \url{https://arxiv.org/abs/2310.11324}.

\bibitem[Yang et~al.(2025)]{Yang2025longbio}
Yijun Yang, Zeyu Huang, Wenhao Zhu, Zihan Qiu, Fei Yuan, Jeff~Z. Pan, and Ivan Titov.
\newblock A controllable examination for long-context language models.
\newblock arXiv:2506.02921, 2025.
\newblock URL \url{https://arxiv.org/abs/2506.02921}.

\bibitem[Zhou et~al.(2023)]{Zhou2023}
Jeffrey Zhou, Tianjian Lu, Swaroop Mishra, Siddhartha Brahma, Sujoy Basu, Yi Luan, Denny Zhou, and Le Hou.
\newblock Instruction-following evaluation for large language models.
\newblock arXiv:2311.07911, November 2023.
\newblock URL \url{https://arxiv.org/abs/2311.07911}.

\end{thebibliography}
\end{document}